\documentclass[a4paper]{article}

\usepackage{INTERSPEECH_v2}
\usepackage{graphicx}
\usepackage{amsmath,bm}
\usepackage{textcomp}
\usepackage{xcolor}
\usepackage{color}
\usepackage{url}
\usepackage{epstopdf}
\usepackage{footnote}
\usepackage{enumitem}
\usepackage{mathtools}
\usepackage{microtype}

\def\vec#1{\ensuremath{\boldsymbol{{#1}}}}

\newcommand\ceil[1]{\lceil#1\rceil}

\title{Multitask Learning with Low-Level Auxiliary Tasks \\ for Encoder-Decoder Based Speech Recognition}
\name{Shubham Toshniwal, Hao Tang, Liang Lu, and Karen Livescu}
\address{
  Toyota Technological Institute at Chicago}
  \email{\{shtoshni, haotang, llu, klivescu\}@ttic.edu}

\begin{document}

\maketitle

\begin{abstract}
End-to-end training of deep learning-based models allows for implicit
learning of intermediate representations based on the final task loss.
However, the end-to-end approach ignores the useful domain knowledge encoded
in explicit intermediate-level supervision.
We hypothesize that using intermediate representations as auxiliary supervision
at lower levels of deep networks may be a good way of combining the advantages of
end-to-end training and more traditional pipeline approaches.  We present
experiments on conversational speech recognition where we use lower-level tasks,
such as phoneme recognition, in a multitask training approach with an
encoder-decoder model for direct character transcription.
We compare multiple types of lower-level tasks and analyze the effects of the auxiliary tasks.
Our results on the Switchboard corpus show that this approach improves
recognition accuracy over a standard encoder-decoder model on the Eval2000 test set.

\end{abstract}

\section{Introduction}
Automatic speech recognition (ASR) has historically been addressed with modular
approaches, in which multiple parts of the system are trained separately.
For example, traditional ASR systems include components like frame classifiers,
phonetic acoustic models, lexicons (which may or may not be learned from data),
and language models~\cite{gales2008application}.
These components typically correspond to different levels of representation,
such as frame-level triphone states, phones, and words.  Breaking up the task
into such modules makes it easy to train each of them separately, possibly on
different data sets, and to study the effect of modifying each component separately.

Over time, ASR research has moved increasingly toward training multiple
components of ASR systems jointly.  Typically, such approaches involve training
initial separate modules, followed by joint fine-tuning using sequence-level
losses~\cite{vesely2013sequence,povey2007evaluation}.
Recently, completely integrated end-to-end training approaches, where all
parameters are learned jointly using a loss at the final output level, have
become viable and popular.  End-to-end training is especially natural for deep
neural network-based models, where the final loss gradient can be backpropagated
through all layers. Typical end-to-end models are based on recurrent neural
 network (RNN) encoder-decoders~\cite{LAS,chorowski15,bahdanau16,liang16} or
connectionist temporal classification (CTC)-based models~\cite{maas_15,zweig2016advances}.

End-to-end training is appealing because it is conceptually simple and allows
all model parameters to contribute to the same final goal, and to do so in the
 context of all other model parameters.  End-to-end approaches have also
 achieved impressive results in ASR~\cite{LAS,zweig2016advances,miao2015eesen}
 as well as other domains~\cite{image_classification_sota,Vinyals,mt_sota}.
 On the other hand, end-to-end training has some drawbacks:  Optimization can be challenging;
the intermediate learned representations are not interpretable, making the
 system hard to debug; and the approach ignores potentially useful
 domain-specific information about intermediate representations, as well as
existing intermediate levels of supervision.

Prior work on analyzing deep end-to-end models has found that different layers
tend to specialize for different sub-tasks, with lower layers focusing on
lower-level tasks and higher ones on higher-level tasks.  This effect has been
found in systems for speech processing~\cite{dbn_layer_learn, nagamine2016role}
as well as computer vision~\cite{zeiler2014visualizing,girshick2014rich}.

We propose an approach for deep neural ASR that aims to maintain the advantages
 of end-to-end approaches, while also including the domain knowledge and
 intermediate supervision used in modular systems.  We use a multitask learning
 approach that combines the final task loss (in our case, log loss on the output
labels) with losses corresponding to lower-level tasks (such as phonetic
recognition) applied on lower layers.  This approach is intended to encapsulate
the intuitive and empirical observation that different layers encode different
levels of information, and to encourage this effect more explicitly.  In other
words, while we want the end-to-end system to take input acoustics and produce
output text, we also believe that at some appropriate intermediate layer, the
network should do a good job at distinguishing more basic units like states or
 phones.  Similarly, while end-to-end training need not require supervision at
 intermediate (state/phone) levels, if they are available then our multitask
 approach can take advantage of them.

We demonstrate this approach on a neural attention-based encoder-decoder
 character-level ASR model.  Our baseline model is inspired by prior
 work~\cite{seq2seq,maas_15,Bahdanau,LAS,liang16}, and our lower-level auxiliary
tasks are based on phonetic recognition and frame-level state classification.
We find that applying an auxiliary loss at an appropriate intermediate layer of
the encoder improves performance over the baseline.

\begin{figure*}[h]
\centering
\includegraphics[width=\textwidth]{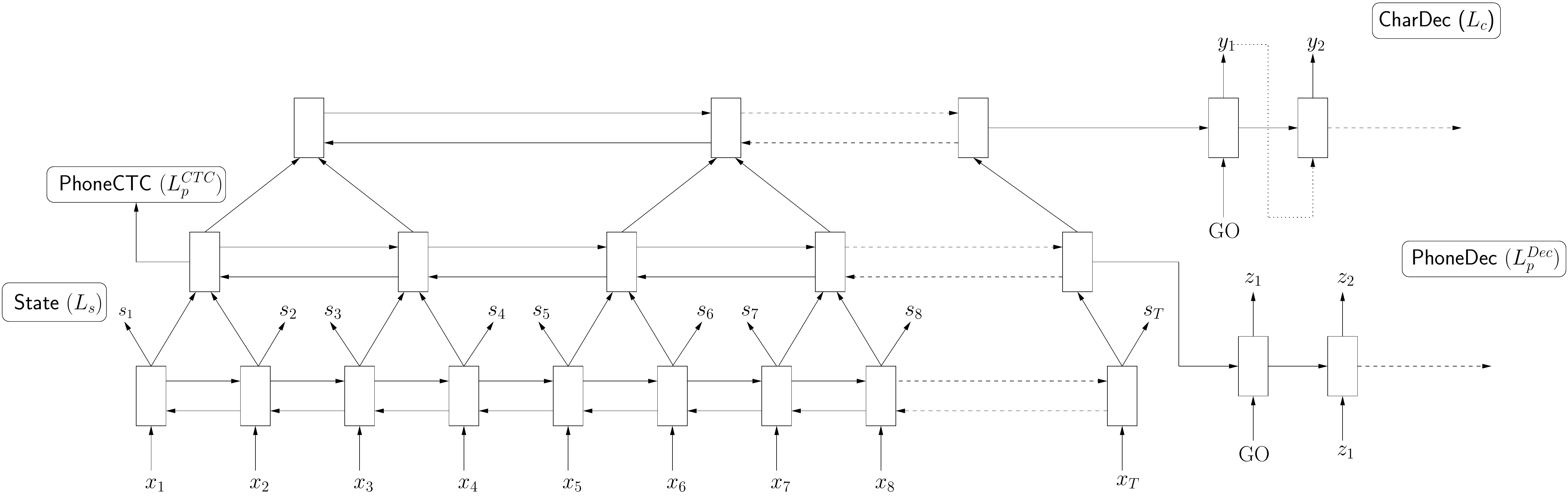}
\caption{Sketch of our training-time model with multiple losses applied at
different layers.  The encoder is a pyramidal bidirectional LSTM
(our experiments use 4 layers; we show 3 layers for simplicity). Different
 hidden state layers of this encoder are used for predicting HMM state label
 $s_i$, phone sequence $\vec{z}$ (using either CTC or a LSTM decoder), and
 finally the output character sequence $\vec{y}$ via a LSTM decoder. The
 dotted line in the character decoder denotes the use of (sampled) model
 predictions~\cite{sampling} during training (for the phone decoder only the
 ground-truth prior phone is used in training). At test time, only the character
decoder is used for transcription.}
\label{fig:pyramid}
\end{figure*}

\section{Related Work}
Multitask training has been studied extensively in the machine learning
literature~\cite{Caruana}.  Its application to deep neural networks has been
successful in a variety of settings in speech and language
processing~\cite{Collobert,mult_luong,cho17,mult_shinji,mult_chan,mult_synthesis}.
 Most prior work combines multiple losses applied at the final output layer of
 the model, such as joint Mandarin character and phonetic recognition
 in~\cite{mult_chan} and joint CTC and attention-based training for English
 ASR~\cite{mult_shinji}.  Our work differs from this prior work in that our
 losses relate to different types of supervision and are applied different
 levels of the model.

The idea of using low-level supervision at lower levels was, to our knowledge,
first introduced by S{\o}gaard \& Goldberg~\cite{mult_gold} for natural language
 processing tasks, and has since been extended by~\cite{mult_soch}. The closest
work to ours is the approach of Rao and Sak~\cite{mult_ctc} using phoneme labels
for training a multi-accent CTC-based ASR system in a multitask setting.  Here
 we study the approach in the context of encoder-decoder models, and we compare
a number of low-level auxiliary losses.  

\section{Models}
\label{sec:model}

The multitask approach we propose can in principle be applied to any type of
deep end-to-end model.  Here we study the approach in the context of
attention-based deep RNNs.  Below we describe the baseline model, followed by
the auxiliary low-level training tasks.

\subsection{Baseline Model}
The model is based on attention-enabled encoder-decoder RNNs, proposed
by~\cite{Bahdanau}. The \emph{speech encoder} reads in acoustic features
$\vec{x} = (\vec{x}_1, \dots, \vec{x}_T)$ and outputs a sequence of high-level
features (hidden states) $\vec{h}$ which the \emph{character decoder} attends to
in generating the output character sequence $\vec{y} = (y_1, \dots, y_{K})$, as
 shown in Figure~\ref{fig:pyramid} (the attention mechanism and a pyramidal
 LSTM layer are \emph{not} shown in the figure for simplicity).

\subsubsection{Speech Encoder}
The speech encoder is a deep pyramidal bidirectional Long Short-Term
Memory~\cite{hochreiter1997lstm} (BiLSTM) network~\cite{LAS}. In the first
layer, a BiLSTM reads in acoustic features $\vec{x}$ and outputs
$\vec{h^{(1)}} = (\vec{h}^{(1)}_1, \dots, \vec{h}^{(1)}_T)$ given by:
\begin{align*}
\overrightarrow{\vec{h}^{(1)}_i} = f^{(1)}(\vec{x}_i, \overrightarrow{\vec{h}^{(1)}_{i - 1}})
& \qquad \overleftarrow{\vec{h}^{(1)}_i} = b^{(1)}(\vec{x}_i, \overleftarrow{\vec{h}^{(1)}_{i + 1}})\\
\vec{h}^{(1)}_i &= (\overrightarrow{\vec{h}^{(1)}_i}; \overleftarrow{\vec{h}^{(1)}_i})
\end{align*}
where $i \in \{1, \dots, T\}$ denotes the index of the timestep; $f^{(1)}(\cdot)$
and $b^{(1)}(\cdot)$ denote the first layer forward and backward LSTMs
respectively\footnote{For brevity we exclude the LSTM equations.  The details
can be found, e.g., in Zaremba {\it et al.} ~\cite{lstm_eqn}.}.

The first layer output $\vec{h^{(1)}} = (\vec{h}^{(1)}_1, \dots, \vec{h}^{(1)}_T)$ is
then processed as follows:
\begin{align*}
  \left.
  \begin{array}{l}
	\overrightarrow{\vec{h}^{(j)}_i} = f^{(j)}([\vec{h}^{(j-1)}_{2i-1};\vec{h}^{(j-1)}_{2i}], \overrightarrow{\vec{h}^{(j)}_{i - 1}})\\
	\overleftarrow{\vec{h}^{(j)}_i} = b^{(j)}([\vec{h}^{(j-1)}_{2i-1};\vec{h}^{(j-1)}_{2i}], \overleftarrow{\vec{h}^{(j)}_{i + 1}})\\
	\vec{h}^{(j)}_i = (\overrightarrow{\vec{h}^{(j)}_i}; \overleftarrow{\vec{h}^{(j)}_i})
  \end{array}
  \right\rbrace \left.
  \begin{array}{r}
  	 \text{for } j = 2, 3, 4
  \end{array}
  \right.
\end{align*}
where $f^{(j)}$ and $b^{(j)}$ denote the forward and backward running LSTMs at layer $j$. Following~\cite{LAS}, we use pyramidal layers to reduces the time resolution of the final state sequence $\vec{h^{(4)}}$ by a factor of $2^{3} = 8$. This reduction brings down the input sequence length, initially $T = |\vec{x}|$, where $|\cdot|$ denotes the length of a sequence of vectors,
close to  the output sequence length\footnote{For Switchboard, the average of number of frames per character is about 7.}, $K = |\vec{y}|$.
For simplicity, we will 
refer to $\vec{h}^{(4)}$ as $\vec{h}$.

\subsubsection{Character Decoder}
The character decoder is a single-layer LSTM that predicts a sequence of
characters $\vec{y}$ as follows:
\begin{equation*}
P(\vec{y}|\vec{x}) = P(\vec{y}|\vec{h}) = \prod_{t=1}^{K}P(y_{t}|\vec{h}, \vec{y_{< t}}).
\end{equation*}
The conditional dependence on the encoder state vectors $\vec{h}$ is represented
 by context vector $\vec{c_{t}}$, which is a function of the current decoder
hidden state and the encoder state sequence:
\begin{align*}
u_{it} &= \vec{v}^\top \text{tanh}(\vec{W_1}\vec{h}_i + \vec{W_2}\vec{d}_t + \vec{b_a}) \\
\vec{\alpha}_{t} & = \text{softmax}(\vec{u}_t) \qquad \vec{c}_t = \sum_{i=1}^{|\vec{h}|} \alpha_{it}\vec{h}_{i}
\end{align*}
where the vectors $\vec{v}, \vec{b_a}$ and the matrices $\vec{W_1}, \vec{W_2}$
are learnable parameters; $\vec{d}_t$ is the hidden state of the decoder at time
 step $t$.
The time complexity of calculating the context vector $\vec{c_{t}}$ for every
time step is $O(|\vec{h}|)$; reducing the resolution on encoder side is crucial
to reducing this runtime.

The hidden state of the decoder, $\vec{d}_{t}$, which captures the previous
character context $\vec{y_{< t}}$, is given by:
$$\vec{d}_{t} = g(\tilde{\vec{y}}_{t-1}, \vec{d}_{t-1}, \vec{c_{t-1}})$$
where $g(\cdot)$ is the transformation of the single-layer LSTM, $\vec{d}_{t-1}$
is the previous hidden state of the decoder, and $\vec{\tilde{y}}_{t-1}$ is a
character embedding vector for $y_{t-1}$, as is typical practice in RNN-based
language models.
Finally, the posterior distribution of the output at time step $t$ is given by:
$$P(y_{t}|\vec{h}, \vec{y_{< t}}) = \text{softmax}(\vec{W_{s}}[\vec{c_{t}}; \vec{d_{t}}] + \vec{b_{s}}),$$
and the character decoder loss function is then defined as
\begin{equation*}
L_c = - \log P(\vec{y} | \vec{x}).
\end{equation*}

\subsection{Low-Level Auxiliary Tasks}
As shown in Figure~\ref{fig:pyramid}, we explore multiple types of auxiliary
tasks in our multitask approach.  We explore two types of auxiliary labels for
multitask learning:  phonemes and sub-phonetic states.  We hypothesize that the
intermediate representations needed for sub-phonetic state classification are
learned at the lowest layers of the encoder, while representations for phonetic
prediction may be learned at a somewhat higher level.
\subsubsection{Phoneme-Based Auxiliary Tasks}
We use phoneme-level supervision obtained from the word-level transcriptions
and pronunciation dictionary. We consider two types of phoneme transcription loss:

{\it Phoneme Decoder Loss:}  Similar to the character decoder described above,
we can attach a phoneme decoder to the speech encoder as well.  The phoneme
decoder has exactly the same mathematical form as the character decoder, but
with a phoneme label vocabulary at the output. Specifically, the phoneme decoder
loss is defined as
\begin{equation*}
L_p^{\text{Dec}} = - \log P(\vec{z} | \vec{x}),
\end{equation*}
where $\vec{z}$ is the target phoneme sequence.
 Since this decoder can be attached at any depth of the four-layer encoder
 described above, we have four depths to choose from.
We attach the phoneme decoder to layer 3 of the speech encoder, and also compare
 this choice to attaching it to layer 4 (the final layer) for comparison with a
 more typical multitask training approach.

{\it CTC Loss:} A CTC~\cite{CTC} output layer can also be added to various
layers of the speech encoder~\cite{mult_ctc}.  This involves adding an extra
softmax output layer on top of the chosen intermediate layer of the encoder, and
applying the CTC loss to the output of this softmax layer. Specifically, let
$\vec{z}$ be the target phoneme sequence, and $k$ be the speech encoder layer
where the loss is applied.
The probability of $\vec{z}$ given the input sequence is
\begin{align*}
P(\vec{z}|\vec{x}) &= \sum_{\pi \in \mathcal{B}^{-1}(\vec{z})} P(\pi | \vec{h}^{(k)})  = \sum_{\pi \in \mathcal{B}^{-1}(\vec{z})} \prod_{j=1}^J P(\pi_j | \vec{h}^{(k)}_j),
\end{align*}
where $\mathcal{B}(\cdot)$ removes repetitive symbols and blank symbols,
$\mathcal{B}^{-1}$ is $\mathcal{B}$'s pre-image,
$J$ is the number of frames at layer $k$
and $P(\pi_j | \vec{h}^{(k)}_j)$ is computed by a softmax function.
The final CTC objective is
\begin{equation*}
L_p^{\text{CTC}} = - \log P(\vec{z} | \vec{x}).
\end{equation*}
The CTC objective computation requires the output length to be less than the
input length, i.e., $|\vec{z}| < J$.  In our case the encoder reduces the time
resolution by a factor of 8 between the input and the top layer, making the top
layer occasionally shorter than the number of phonemes in an utterance.  We
therefore cannot apply this loss to the topmost layer, and use it only at the
third layer.\footnote{In fact, even at the third layer we find occasional
instances (about 10 utterances in our training set) where the hidden state
sequence is shorter than the input sequence, due to sequences of phonemes of
duration less than 4 frames each.
Anecdotally, these examples appear to correspond to incorrect training utterance
alignments}

\subsubsection{State-Level Auxiliary Task}
Sub-phonetic state labels provide another type of low-level supervision that can
 be borrowed from traditional modular HMM-based approaches.  We apply this type
 of supervision at the frame level, as shown in Figure~\ref{fig:pyramid}, using
 state alignments obtained from a standard HMM-based system.
We apply this auxiliary task at layer 2 of the speech encoder.
The probability of a sequence of states $\vec{s}$ is defined as
\begin{equation*}
P(\vec{s} | \vec{x}) = \prod_{m=1}^{M} P(s_m | \vec{x}) = \prod_{m=1}^M P(s_m | \vec{h}^{(2)}_m), 
\end{equation*}
where $P(s_m | \vec{h}^{(2)}_m)$ is computed by a softmax function, and $M$ is
the number of frames at layer 2 (in this case $\ceil{T/2}$).
Since we use this task at layer 2, we subsample the state labels to match the
reduced resolution.
The final state-level loss is
\begin{equation*}
L_s = -\log P(\vec{s} | \vec{x}).
\end{equation*}

\subsubsection{Training Loss}

The final loss function that we minimize is the average of the losses involved. For example,
in the case where we use the character and phoneme decoder losses and the state-level loss,
the loss would be
\vspace{-0.1in}
\begin{equation*}
L = \frac{1}{3} (L_c +  L_p^{\text{Dec}} +  L_s).
\end{equation*}

\section{Experiments}

We use the Switchboard corpus (LDC97S62) \cite{Switchboard}, which contains
roughly 300 hours of conversational telephone speech, as our training set. We
reserve the first 4K utterances as a development set. Since the training set
has a large number of repetitions of short utterances (``yeah'', ``uh-huh",
etc.), we remove duplicates beyond a count threshold of 300. The final training
 set has about 192K utterances. For evaluation, we use the HUB5 Eval2000 data
 set (LDC2002S09), consisting of two subsets: Switchboard (SWB), which is
 similar in style to the training set, and CallHome (CHE), which contains
 unscripted conversations between close friends and family.

For input features, we use 40-dimensional log-mel filterbank features along
with their deltas, normalized with per-speaker mean and variance normalization.
The phoneme labels for the auxiliary task are generated by mapping words to
their canonical pronunciations, using the lexicon in the Kaldi Switchboard
training recipe. The HMM state labels were obtained via forced alignment using
a baseline HMM/DNN hybrid system using the Kaldi NNet1 recipe. The HMM/DNN has
 8396 tied states, which makes the frame-level softmax costly for multitask
 learning.  We use the importance sampling technique described
 in~\cite{large_mt} to reduce this cost.
\vspace{-0.05in}
\subsection{Model Details and Inference}
The speech encoder is a 4-layer pyramidal bidirectional LSTM, resulting in a
8-fold reduction in time resolution.  We use 256 hidden units in each direction
 of each layer. The decoder for all tasks is a single-layer LSTM with 256 hidden
  units.  We represent the decoders' output symbols (both characters and, at
  training time, phonemes) using 256-dimensional embedding vectors. At test time,
  we use a greedy decoder (beam size = 1) to generate the character sequence.
The character with the maximum posterior probability is chosen at every time
step and fed as input into the next time step. The decoder stops after
encountering the ``EOS" (end-of-sentence) symbol.  We use no explicit language model.

We train all models using Adam \cite{adam} with a minibatch size of 64 utterances.
The initial learning rate is 1e-3 and is decayed by a factor of 0.95, whenever
there is an increase in log-likelihood of the development data, calculated after
 every 1K updates, over its previous value. All models are trained for 75K
  gradient updates (about 25 epochs) and early stopping.
To further control overfitting we: (a) use dropout \cite{rnn_dropout} at a rate
 of 0.1 on the output of all LSTM layers (b) sample the previous step's
 prediction \cite{sampling} in the character decoder, with a constant
 probability of 0.1 as in~\cite{LAS}. 

\vspace{-0.1in}
\subsection{Results}
\vspace{-0.03in}
We evaluate performance using word error rate (WER). We report results on the
 combined Eval2000 test set as well as separately on the SWB and CHE subsets.
 We also report character error rates (CER) on the development set.

\begin{table}[t]
\footnotesize
\centering
  \caption{Character error rate (CER, \%) and word error rate (WER, \%)
  results on development data.}
  \vspace{-0.1in}
\begin{tabular}{ l  c  c }
\hline
\textbf{Model} & \textbf{Dev CER } &\textbf{Dev WER}\\\hline
Enc-Dec (baseline) &  14.6 & 26.0 \\
Enc-Dec + PhoneDec-3  &  13.8 & 24.9\\
Enc-Dec + PhoneDec-4 & 14.5  & 25.9 \\
Enc-Dec + PhoneCTC-3 & 14.0  & 25.3 \\
Enc-Dec + State-2 & 13.6 &  24.1 \\\hline
\textbf{Enc-Dec + PhoneDec-3 + State-2} & \textbf{13.4} &  \textbf{24.1} \\\hline
\end{tabular}
\label{tab:dev_res}
\vskip-2mm
\end{table}

\begin{figure}[t]
\centering
\includegraphics[width=0.4\textwidth, height=150pt]{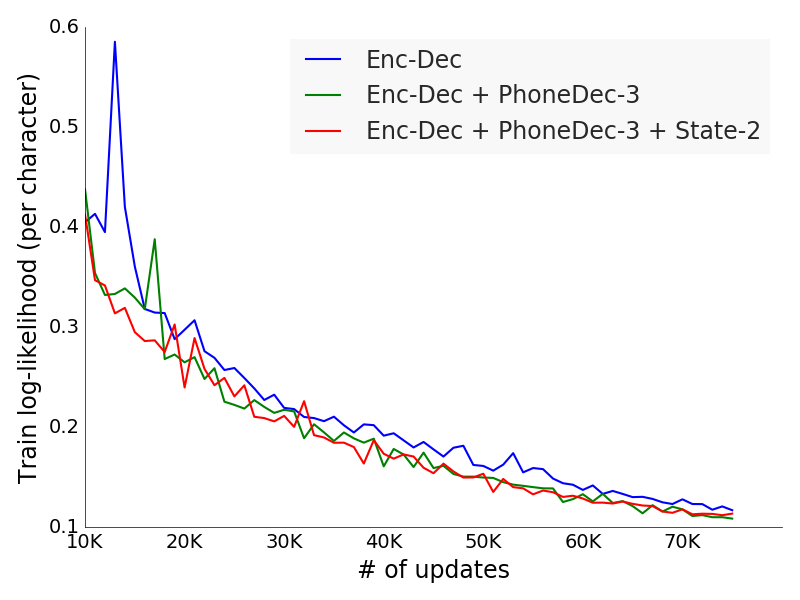}
\vspace{-0.1in}
\caption{Log-likelihood of train data (per character) for different model variations.}
\label{fig:train_likelihood}
\vskip-2mm
\end{figure}

Development set results are shown in Table~\ref{tab:dev_res}.  We refer to the
 baseline model as ``Enc-Dec" and the models with multitask training as
 ``Enc-Dec + [auxiliary task]-[layer]". Adding phoneme recognition as an
 auxiliary task at layer 3, either with a separate LSTM decoder or with CTC,
 reduces both the character error rates and the final word error rates.

In order to determine whether the improved performance is a basic multitask
training effect or is specific to the low-level application of the loss, we
 compare these results to those of adding the phoneme decoder at the topmost
 layer (Enc-Dec + PhoneDec-4). The top-layer application of the phoneme loss
 produces worse performance than having the supervision at the lower (third)
 layer.  Finally, we obtain the best results by adding both phoneme decoder
 supervision at the third layer and frame-level state supervision at the second
  layer (Enc-Dec + PhoneDec-3 + State-2). The results support the hypothesis
  that lower-level supervision is best provided at lower layers.
Table~\ref{tab:test_res} provides test set results, showing the same pattern of
 improvement on both the SWB and CHE subsets.
For comparison, we also include a variety of other recent results with neural
end-to-end approaches on this task.
Our baseline model has better performance than the most similar previous
encoder-decoder result~\cite{liang16}.  With the addition of the low-level
auxiliary task training, our models are competitive with all of the previous
end-to-end systems that do not use a language model.

Figure~\ref{fig:train_likelihood} shows the training set log-likelihood for the
 baseline model and two multitask variants. The plot suggests that multitask
 training helps with optimization (improving the training error).  Training
 error is very similar for both multitask models, while the development set
 performance is better for one of them (see Table~\ref{tab:dev_res}), suggesting
  that there may also be an improved generalization effect and not only improved
  optimization.

\begin{table}[t]
\footnotesize
\centering
\caption{WER (\%) on Eval2000 for different end-to-end models. PhoneDec-$n$ refers to a phoneme decoder applied at layer $n$ of the encoder. Similarly, PhoneCTC-3 means phoneme CTC loss applied at layer 3 and State-2 means state-label supervision applied at layer 2 of the encoder. 
}
\vspace{-0.1in}
\begin{tabular}{ l  c  c  c }
\hline
\textbf{Model} & \textbf{SWB} & \textbf{CHE} & \textbf{Full}\\\hline
Our models\\
\hspace{0.1in}Enc-Dec (baseline) & 25.0 & 42.4 & 33.7 \\
\hspace{0.1in}Enc-Dec + PhoneDec-3 & 24.5 & 40.6 & 32.6 \\
\hspace{0.1in}Enc-Dec + PhoneDec-4 & 25.4 & 41.9 & 33.7 \\
\hspace{0.1in}Enc-Dec + PhoneCTC-3 & 24.6 & 41.3 & 33.0 \\
\hspace{0.1in}Enc-Dec + State-2 &  24.7 & 42.0 & 33.4 \\\hline
\hspace{0.1in}\textbf{Enc-Dec + PhoneDec-3 + State-2} &  \textbf{23.1} & \textbf{40.8} & \textbf{32.0} \\\hline
Lu et al.\cite{liang16} &  &  & \\
\hspace{0.1in} Enc-Dec & 27.3 & 48.2 & 37.8 \\
\hspace{0.1in} Enc-Dec (word) + 3-gram &  25.8 & 46.0 & 36.0 \\\hline
Maas et al.~\cite{maas_15} & & & \\
\hspace{0.1in}CTC & 38.0 & 56.1 & 47.1\\
\hspace{0.1in}CTC + 3-layer RNN LM & 21.4 & 40.2 & 30.8\\\hline
Zweig et al.~\cite{zweig2016advances} & & & \\
\hspace{0.1in}Iterated CTC & 24.7 & 37.1 & ---\\
\hspace{0.1in}CTC + Char Ngram & 19.8 & 32.1 & --- \\
\hspace{0.1in}CTC + Dictionary + Word Ngram & 14.0& 25.3 & --- \\\hline
\end{tabular}
\label{tab:test_res}
\vskip-3mm
\end{table}

\vspace{-0.12in}
\section{Conclusion}
\vspace{-0.05in}
We have presented a multitask training approach for deep end-to-end ASR models
in which lower-level task losses are applied at lower levels, and we have
explored this approach in the context of attention-based encoder-decoder models.
  Results on Switchboard and CallHome show consistent improvements over baseline
  attention-based models and support the hypothesis that lower-level supervision
  is more effective when applied at lower layers of the deep model.  We have
compared several types of auxiliary tasks, obtaining the best performance with
a combination of a phoneme decoder and frame-level state loss.  Analysis of
model training and performance suggests that the addition of auxiliary tasks
can help in either optimization or generalization.

Future work includes studying a broader range of auxiliary tasks and model
 configurations.  For example, it would be interesting to study even deeper
models and word-level output, which would allow for more options of intermediate
tasks and placements of the auxiliary losses.  Viewing the approach more
 broadly, it may be fruitful to also consider higher-level task supervision,
 incorporating syntactic or semantic labels, and to view the ASR output as an
intermediate output in a more general hierarchy of tasks.

\vspace{-0.1in}
\section{Acknowledgements}
\vspace{-0.06in}
{\footnotesize We are grateful to William Chan for helpful discussions, and to the speech group at TTIC, especially Shane Settle, Herman Kamper, Qingming Tang, and Bowen Shi for sharing their data processing code.  This research was supported by a Google faculty research award.}
\bibliographystyle{IEEEtran}
\bibliography{asr}

\end{document}